\documentclass[twoside,leqno,twocolumn]{article}

\usepackage[letterpaper]{geometry}
\usepackage{ltexpprt}
\usepackage{color,soul}
\usepackage{cite}
\usepackage{amsmath,amssymb,amsfonts}
\usepackage{graphicx}
\usepackage{textcomp}
\usepackage{xcolor}
\usepackage{algorithm,algcompatible,amsmath}

\usepackage{mathtools}
\usepackage{thmtools} 
\usepackage{booktabs}
\usepackage{hyperref}

\def\BibTeX{{\rm B\kern-.05em{\sc i\kern-.025em b}\kern-.08em
    T\kern-.1667em\lower.7ex\hbox{E}\kern-.125emX}}
\declaretheorem{definition} 

\begin{document}

\title{\Large Joint Time Series Chain: Detecting Unusual Evolving Trend across Time Series}
\author{Li Zhang\thanks{
Department of Computer Science, George Mason University, \{lzhang18, jessica\}@gmu.edu}\qquad Nital Patel \thanks{
TD Analytics and Technology Automation, Intel Corporation, \{nital.s.patel, xiuqi.li\}@intel.com}\qquad Xiuqi Li\footnotemark[2] \qquad Jessica Lin\footnotemark[1]}
\date{}

\maketitle

\fancyfoot[R]{\scriptsize{Copyright \textcopyright\ 2022 by SIAM \\
Unauthorized reproduction of this article is prohibited}}

\begin{abstract} \small\baselineskip=9pt
Time series chain (TSC) is a recently introduced concept that captures the evolving patterns in large scale time series. Informally, a time series chain is a temporally ordered set of subsequences, in which consecutive subsequences in the chain are similar to one another, but the last and the first subsequences maybe be dissimilar. Time series chain has the great potential to reveal latent unusual evolving trend in the time series, or identify precursor of important events in a complex system. Unfortunately, existing definitions of time series chains only consider finding chains in a single time series. As a result, they are likely to miss unexpected evolving patterns in interrupted time series, or across two related time series. To address this limitation, in this work, we introduce a new definition called \textit{Joint Time Series Chain}, which is specially designed for the task of finding unexpected evolving trend across interrupted time series or two related time series. Our definition focuses on mitigating the robustness issues caused by the gap or interruption in the time series. We further propose an effective ranking criterion to identify the best chain. We demonstrate that our proposed approach outperforms existing TSC work in locating unusual evolving patterns through extensive empirical evaluations. We further demonstrate the utility of our work with a real-life manufacturing application from Intel. Our source code is publicly available at the supporting page (\url{https://github.com/lizhang-ts/JointTSC}).

\end{abstract}

\section{INTRODUCTION}

Time Series pattern mining has attracted significant amount of attentions from researchers for its broad applications in different domains~\cite{yeh2016matrix,gao2017trajviz,zhang2020semantic}. Recently, time series chain (TSC) is introduced to capture the change of patterns over time~\cite{zhu2017tsc, zhu2019introducing,imamura2020matrix}. Informally, time series chain is an ordered set of time series subsequences extracted from a single time series, in which the adjacent subsequences in the chain are similar, but the beginning and the end subsequences might diverge from each other~\cite{zhu2017tsc}. Intuitively, a time series chain is uniquely designed to capture any potential \textit{evolving drift} accumulated over time, which widely exists in many natural phenomena and mechanisms such as animal behavior~\cite{zhu2019introducing}, environment change, human metabolization~\cite{zhu2017tsc} and mechanical equipment worn-out~\cite{bloch1997practical}. For example, in manufacturing process such as wielding, it is often important to monitor the change in temperature, as maintaining a steady temperature of the machine is critical to ensure that all materials and parts are processed under uniform conditions. As time goes by, the system is no longer stable due to changes induced by the heat generated during the process, and it takes longer and is harder for the system to recover back to the setpoint. These changes in temperature affect the yield of the product. Once the system becomes unstable, certain maintenance and cleaning process will need to take place. Time series chain could potentially capture the evolving drift that indicates the latent status change in the system, and it can be useful for regular checking of the sensor status or ad hoc maintenance.

Unfortunately, all current time series chain definitions have been focusing on finding TSCs in a single time series. That means the data needs to be collected continuously without any interruption. However, such data requirement, in practice, can be too strict and unrealistic for many applications. First, the data itself might naturally be collected with interruptions. For example, in cardiac diagnosis, a patient may undergo continuous cardiac monitoring using a standard bedside monitor for 24 hours, released, and then put back on the monitor later when more serious symptoms occur~\cite{adami2019electrocardiographic}. Second, even if all the data were collected continuously without interruption, in many industry settings such as oil pipeline maintenance, Jardine et al.~\cite{jardine2006review} pointed out that the continuous flow of data from the beginning to the end might introduce additional noises, which may contaminate the data with inaccurate information. 

To simultaneously address the data gap and noise issues, we consider a new problem of finding a single time series chain \textit{across} two (separate but related) time series. The detected chain should identify unexpected or unusual trend from one time series to the other. In practice, the two time series can be two segments of data interrupted by some contiguous period of noises, or simply two time series collected from the same source. To better identify unusual trend, we hereby assume that one of the time series is relatively stable (normal) compared to the other time series. For convenience, we refer to the stable time series as the \textbf{reference time series} and the time series of interest as \textbf{target time series}. The two time series do not have to have the same length.
\begin{figure}[ht]
    \centering
    \includegraphics[width=85mm]{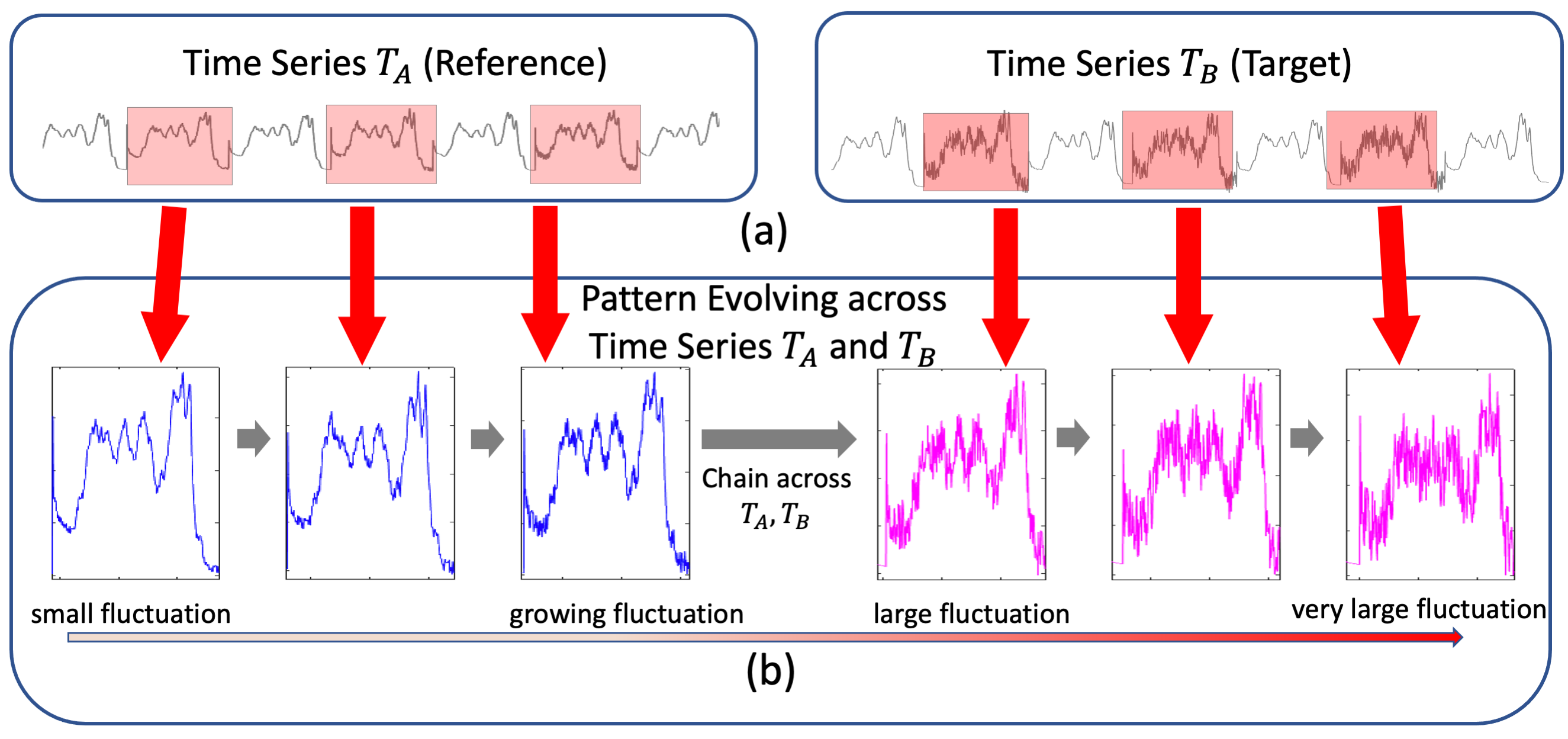}
    \vspace{-2.5mm}
    \caption{An illustration of our problem definition of Joint Time Series Chain (JTSC). (a) Two time series, $T_A$ and $T_B$ are shown. (b) JTSC captures evolving patterns across two time series. }
    
    \label{fig:intro_new}
    \vspace{-5mm}
\end{figure}

Specifically, in this work, we introduce a new time series chain definition named \textbf{Joint Time Series Chain (JTSC)}, which is specially designed to identify unusual evolving trend in the target time series given a reference time series.
We illustrate our definition with an example in Figure 1. 

Figure 1(a) shows two time series $T_A$ (reference time series) and $T_B$ (target time series). Figure 1(b) shows one example of Joint Time Series Chain. Its original position is highlighted in orange boxes in Figure 1(a). From the figure, it is easy to observe that in $T_A$, the chain is relatively stable, and the subsequences in the chain are similar with only a little evolving local fluctuations. However, while the first subsequence in the chain in $T_B$ is similar to those in $T_A$, the rest of the subsequences capture more vigorous evolving fluctuations that do not exist in $T_A$. In practice, this kind of evolution reflects a less stable system, and could be a good indication of precursors of certain events such as mechanical failures. 

Detecting such an evolving chain is a non-trivial task. One may argue that simply concatenating the two time series, and applying existing TSC methods on the concatenated time series could potentially identify an overall chain that captures the unexpected evolving trend. However, such an approach will not work, as existing TSC methods are designed to capture steady trend with small changes. When the chain detected in the reference part of the time series propagates to the target time series, the more abrupt changes in behavior in the target time series would cause the algorithm to consider it a different chain. Therefore, existing TSC definitions are not suitable to handle our problem.

We would like to point out that joint time series chain should not be confused with other time series pattern mining primitives, such as contract set~\cite{lin2006group}, motif~\cite{chiu2003probabilistic}, or time series discord~\cite{keogh2005hot}. Specifically, we are interested in identifying a sequence of subsequences that reflect an evolution of (unusual) behavior, as opposed to finding a single subsequence indicating such behavior.
We summarize our contributions as follows: 
\vspace{-1mm}
\begin{itemize}
\itemsep0em 
    \item We propose a new definition named \textit{Joint Time Series Chain (JTSC)} for finding trends across two related but disjoint time series. The proposed definition mitigates the issues caused by the gap between the time series.
    \item We propose a novel ranking criterion that considers the deviation between the two time series. This allows us to identify unusual evolving trends in a targeted time series. 
    \item We demonstrate that our JTSC algorithm significantly outperforms the state-of-the-art TSC in locating unusual evolving trends via extensive experiments. We further demonstrate that JTSC can be useful for monitoring of the sensor status for regular and ad hoc maintenance in a manufacturing application at Intel.
\end{itemize}
\vspace{-2mm}
The paper is organized as follows. In Section 2, we briefly review the related work and provide the background information. In Section 3, we provide the fundamental definitions on time series chain. In Section 4, we discuss the limitations of existing TSC definitions in our problem. 
In Section 5, we provide details for our design of Joint Time Series Chain and present the proposed algorithm. In Section 6, we propose a ranking criterion to select the top chain. In Section 7, we perform extensive experiments to show that our method outperforms the state-of-the-art methods. We also present a case study with real world Intel manufacturing application. Finally, we conclude this paper and discuss the future work in Section 8. 
\vspace{-3mm}
\section{BACKGROUND AND RELATED WORK}
We discuss the related work for time series chain in this section. Our discussion is relatively brief since there has not been much similar work. The most closely related work, by Zhu et al~\cite{zhu2017tsc}, introduced the concept of time series chain. It is the first work focusing on capturing the evolving shape in time series research domain. They define a time series chain as the longest sequence of evolving motifs (approximately repeating patterns), and enforce every adjacent pair of subsequences in the chain to be a nearest neighbor to each other in both left and right directions. Although the concept is very intuitive, the chains are not very robust against noise, and the bi-directional nearest neighbor condition could be too strong for complex data. Imamura et al.~\cite{imamura2020matrix} relax the bi-directional condition, and only use left nearest neighbors to find backward chains. We will elaborate on the limitations of these two chain definitions in Section 4. Nevertheless, neither of these two methods considers looking \textit{across} two related time series to capture unusual change of behavior from the reference time series to the target time series. More concretely, since the original TSC algorithms are designed to track behavior that is mostly consistent, with slight deviation over time, if some interruption occurs in the time series and disrupts the behavior in the time series, the chain will terminate. Our Joint Time Series Chain is motivated by real-life applications and mitigates the issues caused by the gap between the time series.

There is also other work that focuses on speeding up the original time series chain technique \cite{wang2019discovering}. Since they use the same TSC definition, we exclude these works from our experimental comparison. In the experiments, we compare with the original time series chain~\cite{zhu2017tsc} and backward time series chain~\cite{imamura2020matrix} techniques. We will refer to them as TSC17 and TSC20, respectively.  

Note that this work is unrelated to the task of ``time series join"  \cite{mueen2014time, vinh2016novel}, which aims to find their most correlated segments among two time series.
\vspace{-2mm}
\section{DEFINITIONS}
We begin with fundamental definitions of time series. \vspace{-2mm}
\begin{definition}
A \textbf{Time series} $T = [t_1, t_2, \ldots, t_n]$ is an ordered list of data points, where $t_i$ is a finite real number and $n$ is the length of time series $ T$.
\vspace{-2mm}
\end{definition}

\begin{definition}
A \textbf{time series subsequence} $S^T_i= [t_i, t_{i+1}, \ldots, t_{i+l-1}]$ is a contiguous set of points in time series $ T$ starting from position $i$ with length $l$. Typically $l\ll n$, and $1\leq i\leq n-l+1$.
\vspace{-2mm}
\end{definition}
For simplicity, if we only refer to a subsequence of one time series, we use $S_i$ interchangeably with $S^T_i$. 
\noindent Subsequences can be extracted from time series $T$ by sliding a fixed-length window across the time series. 

Given two subsequences of the same length, the Euclidean Distance is often used to measure their differences. To achieve scale and offset invariance, each subsequence must be properly normalized before the actual distance computation. Z-normalized Euclidean Distance is defined below.
\vspace{-1mm}
\begin{definition}
A \textbf{z-normalized Euclidean Distance} $d(S_p, S_q)$ of subsequences $S_p$, $S_q$ of length $l$ is computed as 
\vspace{-3mm}
$$\tiny {d(S_p,S_q)=\sqrt{\sum_{m = 1}^{l}{(\frac{t_{p+m-1} - \mu_p}{\sigma_p} - \frac{t_{q+m-1} - \mu_q}{\sigma_q} })^2},}$$
where $\mu_p$, $\sigma_p$ and $\mu_q$, $\sigma_q$ are the means and standard deviations of subsequences $S_p$ and $S_q$ respectively. 

\end{definition}

We next describe definition of distance profile~\cite{yeh2016matrix} based on z-normalized Euclidean Distance:

\begin{definition}
\textbf{distance profile} Given a query subsequence $Q$ and a time series $T$, a distance profile $D_{Q,T}$ is a vector containing the Euclidean distances between $Q$ and each subsequence of the same length in time series $T$. Formally, $D_{Q,T} = [d(Q,S^T_1), d(Q,S^T_2), \cdots, d(Q,S^T_{n-l+1})] \quad$. In a special case when the query subsequence $Q \in T$, for simplicity, we simply denote it as $D_i$.
\end{definition} 
\vspace{-2mm}
Before introducing the time series chain, we state the definitions of left/right distance profiles~\cite{zhu2017tsc} and matrix profiles~\cite{yeh2016matrix}. 
\vspace{-2mm}
\begin{definition}
A \textbf{left distance profile}, $DL_i$, of time series $T$ is a vector containing the Euclidean distances between a given subsequence $S_i \in T$ and every subsequence to the left of $S_i$. Formally, $DL_i = [d(S_i,S_1), d(S_i,S_2), \cdots, d(S_i,S_{i-l)}].$

\end{definition}

\begin{definition}
\textbf{right distance profile} $DR_i$ is a vector where: $DR_i = [d(S_i,S_{i+l}),\cdots, d(S_i,S_{n-l+1})],$
\end{definition}
Note that, for all distance profiles, we exclude any subsequence that overlaps with the query subsequence to ensure that all comparisons are meaningful and non-trivial.

\noindent Based on distance profile, we can define left/right/joint matrix profiles~\cite{yeh2016matrix, gharghabi2018matrix}.

\begin{definition}
A \textbf{left matrix profile} $MP_L$ is a two dimensional vector of size 2-by-$(n-l+1)$ where $MP_L(1,i)=min(DL_i)$ and $MP_L(2,i)=argmin(DL_i)$ 
\end{definition}

\begin{definition}
\textbf{right matrix profile} is a two dimensional vector of size 2-by-$(n-l+1)$ such that $MP_R(1,i)=min(DR_i),MP_R(2,i)=argmin(DR_i)$
\end{definition}

\noindent where $MP_L(1,i)$ and $MP_R(1,i)$ store the distances between $S_i$ and the most similar subsequence before and after $i$, respectively. $MP_L(2,i)$ and $MP_R(2,i)$ store the indices of such subsequences. For the rest of the paper, we also refer to such similar subsequences as \textbf{\textit{left nearest neighbor}} and \textbf{\textit{right nearest neighbor}}, respectively. 

We describe the join matrix profile or AB/BA join matrix profile~\cite{gharghabi2018matrix}, a concept which will be used in our joint time series chain:
\vspace{-1mm}
\begin{definition}
\textbf{AB/BA Join Matrix Profile} Given two time series $T_A$ and $T_B$, an AB Join Matrix Profile, $MP_{AB}$, is a matrix profile such that $MP_{AB}(1,i)=min(D_{S^A_i,T_B})$ and $MP_{AB}(2,i)=argmin(D_{S^A_i,T_B})$. Analogously, a BA Join Matrix Profile $MP_{BA}$ is defined as: $MP_{BA}(1,j)=min(D_{S^B_j,T_A})$ and $MP_{BA}(2,j)=argmin(D_{S^B_j,T_A})$. 
\end{definition}

Now we will describe two existing time series chain definitions. We follow Imamura et al.~\cite{imamura2020matrix} and refer to them as TSC17 and TSC20, respectively. 
Zhu et al. (TSC17)~\cite{zhu2017tsc}  define a bi-directional time series chain as follows: 
\vspace{-2mm}
\begin{definition}
A \textbf{bi-directional time series chain (TSC17)} of time series $T$ is a finite ordered set of time series subsequences. $TSC17 =[S_{C_1}, S_{C_2}, S_{C_3}, \cdots, S_{C_m}]$ where $C_1 < C_2 <\cdots < C_m$ are indices in time series $T$, such that $MP_R(2,MP_L(2,C_{i})) = C_{i}$, and $m$ is the length of time series chain. 
\end{definition}
Specifically, any node $S_{C_i}$ in $TSC17$ is the right nearest neighbor of $S_{C_{i-1}}$, and the left nearest neighbor of $S_{C_{i+1}}$.

Imamura et al. (TSC20)~\cite{imamura2020matrix} relax the constraint and use left nearest neighbor to define time series chain. They define the time series chain as follows:
\vspace{-2mm}
\begin{definition}
A \textbf{backward time series chain} of time series $T$ is a finite ordered set of time series subsequences and  $TSC_{BWD} =[S_{C_1}, S_{C_2}, S_{C_3}, \cdots, S_{C_m}]$ where $C_1 < C_2 <\cdots < C_m$ are indices in time series $T$, such that $MP_L(2,S_{C_i+1}) = C_{i}$. 
\end{definition}
\vspace{-1mm}

Similarly, we define the forward time series chain as we will use it later. 
\vspace{-1mm}
\begin{definition}
A \textbf{forward time series chain} of time series $T$ is a finite ordered set of time series subsequences and  $TSC_{FWD} =[S_{C_1}, S_{C_2}, S_{C_3}, \cdots, S_{C_m}]$ where $C_1 < C_2 <\cdots < C_m$ are indices in time series $T$, such that $MP_R(2,S_{C_i}) = C_{i+1}$. 
\end{definition}
For brevity, we refer to a subsequence in time series chain as a \textbf{node}. In addition, for backward and forward chains, we define the \textbf{start node} and \textbf{end node} based on their search direction in the chain. Specifically, for a forward chain, $S_{C_1}$ is called the start node and $S_{C_m}$ is called the end node. For a backward chain, $S_{C_m}$ is the start node and $S_{C_1}$ is the end node.

Before introducing our definition of Joint Time Series Chain, we discuss the limitations of previous TSCs.  
\vspace{-3mm}

\section{Limitations of Previous TSCs on Our Task}

As mentioned in Introduction, one potential solution to identify a time series chain across two time series is to concatenate them and apply TSC17 or TSC20 on the concatenated time series. Here we explain why this approach does not work for our problem.
\vspace{-2mm}
\subsection{TSC17}
The original definition of time series chain (TSC17) is unsuitable for our task due to the following reasons:

\noindent\textbf{Short Chain Length} The discovered chain often consists of very few nodes. As the authors of TSC20 \cite{imamura2020matrix} pointed out, TSC17 is susceptible to noises. As a result, TSC17 tends to identify short chains, or fail to identify apparent chains when there is noise in the data. 

\noindent\textbf{Chain in Single Time series} In our problem setting where one might observe deviating behavior from the reference time series to the target time series, the nearest neighbor of a subsequence would likely be in the same time series rather than crossing over to the other one. As a result, the detected chain would likely be congregating on only one time series and unable to detect the deviation.
\vspace{-2mm}
\subsection{TSC20 -- Imbalanced Chain}
Compared to TSC17, TSC20~\cite{imamura2020matrix} relaxes the strong constraint on adjacent nodes to include more nodes in the chain, and make it more robust in the presence of noise. Although TSC20 can detect longer chains compared to TSC17, the chain is likely to gravitate towards one side of the time series, as it relies on backward chain. 
\vspace{-2mm}
\begin{figure}[h]
    \centering
    \includegraphics[width=85mm]{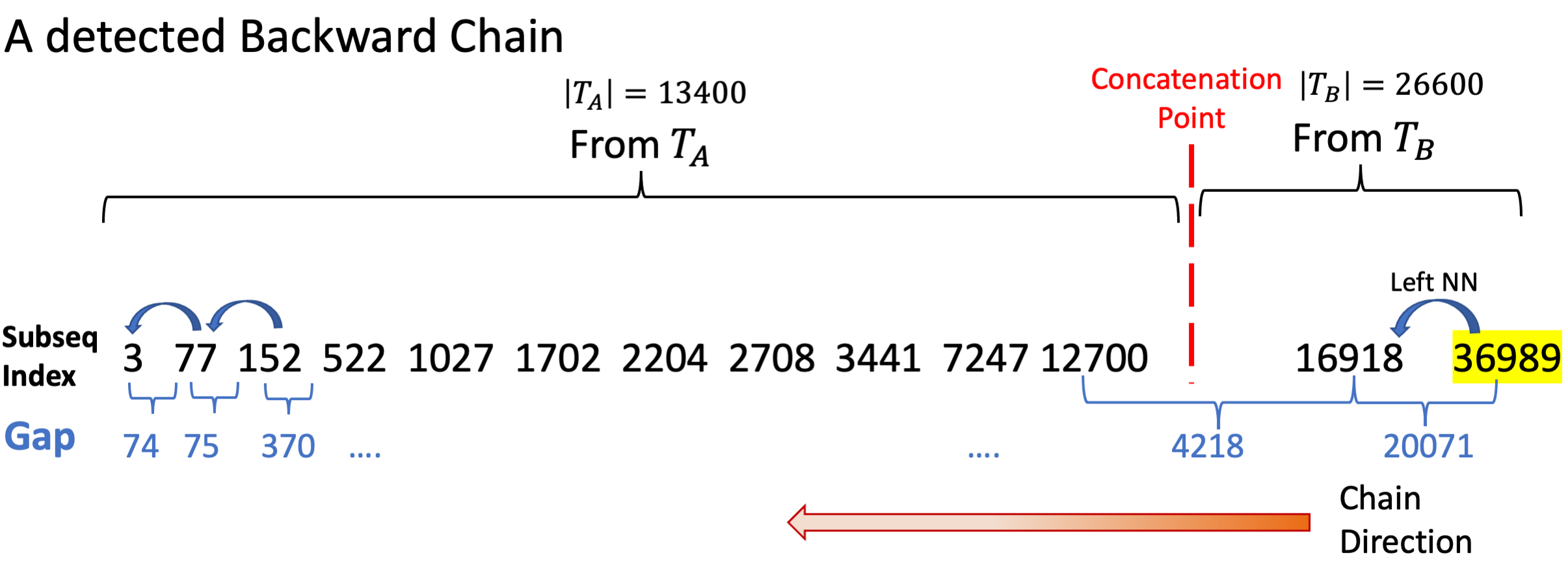}
    \caption{An example of a detected backward time series chain on the concatenated reference and target time series.}
    \vspace{-2mm}
    \label{fig:backward_limitation}
    \vspace{-1mm}
\end{figure}
\vspace{-1mm}
We demonstrate this problem of one directional chain with a simple example using a real dataset from Intel in Figure~\ref{fig:backward_limitation}. $T_A$ and $T_B$, each collected before and after some interruption occurred, respectively, are concatenated to form one long time series. In the figure, the top row (in black) shows the indices of the nodes in the chain, and the bottom row (in blue) shows the intervals between adjacent nodes. It can be seen that the intervals decrease sharply as the chain propagates towards the beginning of $T_A$. For example, the interval between the rightmost node at index 36989 and its left neighbor at 16918 has more than 20000 points. The first two leftmost intervals have 74 and 75 points, respectively, which are less than 0.2\% of the time series length. It is easy to see that the indices get denser and denser towards the left, since fewer subsequence candidates become available to choose from as the backward chain propagates. Although TSC20 uses some filtering strategy to remove the unwanted nodes towards the end of the backward chain, it is still likely to find nodes mostly in one time series (the left, reference time series) rather than in both time series. 

If we use a forward chain, we would encounter a similar problem, as most of the nodes would be located in the right time series. 
\vspace{-2mm}
\section{Joint Time Series Chain}
Being aware of the above limitations, we introduce a new definition named Joint Time Series Chain (JTSC), which is specially designed for identifying unusual evolving trend across two time series. 

Given a reference time series $T_A$ and a target $T_B$, we would like to find a Joint Time Series Chain (JTSC) that consists of a backward TSC in $T_A$ and a forward TSC in $T_B$. In addition, the last node in $T_A$ and the first node in $T_B$ are nearest neighbors in the other time series through AB/BA join matrix profile. 
Formally, Joint Time Series Chain is defined as:
\vspace{-2mm}
\begin{definition}
\textbf{Joint Time Series Chain} Given a reference time series $T_A$ and a target $T_B$ of length $n_A$ and $n_B$ respectively, a Joint Time Series Chain is a single chain $JTSC(T_A,T_B) =[S^A_{C_1}, \cdots, S^A_{C_{m_A}}, S^B_{C'_1}\cdots, S^B_{C'_{m_B}}]$ that satisfies all of the following conditions:
\begin{itemize}
    \item $[S^A_{C_1}, \cdots, S^A_{C_{m_A}}]$ is a $TSC_{BWD}$ in $T_A$; that is, $MP_L(2,S_{C_j+1})=C_{j}$;
    \item $[S^B_{C'_1}\cdots, S^B_{C'_{m_B}}]$ is a $TSC_{FWD}$ in $T_B$; that is, $MP_R(2,S_{C'_j})=C'_{j+1}$;
    \item $MP_{BA}(2,MP_{AB}(2,S^A_{C_{m_A}})) = C_{m_A}$.
\end{itemize}

\end{definition}
\noindent where $(S_{C_{m_A}}, S_{C'_{1}})$ are called \textbf{junction nodes} for a time series chain $JTSC_{AB}$. The backward chain $[S_{C_1}, \cdots, S_{C_{m_A}}]$ in $T_A$ is called \textbf{reference sub-chain} of $JTSC_{AB}$. Similarly, the forward chain $[S_{C'_1}\cdots, S_{C'_{m_B}}] $ in $T_B$ is called \textbf{target sub-chain}.

We will describe the intuition for such a design and our contributions in the next few subsections.
\vspace{-2mm}
\subsection{Advantage of Joint Time Series Chain} 
Joint Time Series Chain can identify more balanced chain compared to TSC17 and TSC20. We can attribute the result to (1) the bi-directional join via the junction nodes, and (2) connecting two independent, single-directional chains within the time series. 
\vspace{-2mm}
\subsubsection{Importance of Junction Nodes} Since JTSC relies on junction nodes to connect across two time series, the quality of junction nodes is of critical importance that will ultimately determine the quality of the final JTSC. If either of the junction nodes is a noisy and less meaningful node, the resulting JTSC will consequently become a meaningless chain. 
\vspace{-2mm}
\subsubsection{Importance of Bi-directional Connection for Junction Nodes} Our design of JTSC is largely motivated by improving the quality of junction nodes. First, as we mentioned earlier in Section 4.2, the quality of the one-directional chain (e.g., backward chain) deteriorates as the chain propagates. However, the ideal positions of the junction nodes are located close to the end of $T_A$ and the beginning of $T_B$, respectively, in order to keep the most potential subsequences for a longer chain. Thus, it would be best to use those locations as the start nodes, rather than the end nodes, for chain search. 

We enforce a strong constraint -- the junction nodes need to be the nearest neighbors of one another, making it unlikely for a noise node to become a junction node. The independent bi-directional searches and joins would ensure that the junction nodes are of high quality, and thus result in better JTSC. 
\vspace{-1mm}
\section{Ranking Score for Joint Time Series Chain}
We next introduce a ranking score to identify the most interesting JTSC. Intuitively, an interesting chain should reflect an evolving trend across two time series that can identify unusual evolving behavior that may be useful as precursor of some adversarial event. 
\vspace{-2mm}
\subsection{Ranking Score}
For simplicity, we denote JTSC for $T_A$ and $T_B$ as $JTSC_{AB}(T_A, T_B) = [CA_1, CA_{2}, \cdots, CA{m_A}, CB_{1}, \cdots CB{m_B}]$, where $CA_{i}\in T_A$, and $CB_{j}\in T_B$. The overall ranking score is 
\begin{equation}
    Score_{rank} =  \frac{Score_{Ref}+1}{Score_{Fluc}+1}\cdot Score_{Evolve},
\end{equation}
where a Laplace smoothing is applied on the two-score ratio. $Score_{Fluc}$, $Score_{Evolve}$, and  $Score_{Ref}$ are the metrics to evaluate the fluctuation, evolving capability, and deviation of the chain from the reference time series, respectively. We will elaborate on these three score components next.  
\vspace{-2mm}
\subsection*{Fluctuation Score}
To evaluate the anomalousness of an evolving trend, the evolving change in $T_A$ should be relatively stable since we will use it as a reference for the normal behavior of the system. Thus, the adjacent nodes in the desired chain in $T_A$ should be relatively similar and steady. Moreover, since we are only interested in the evolving chain across two time series, the junction nodes between $T_A$ and $T_B$ need to be similar. Based on these intuitions, we design the fluctuation score as follows:
 
\begin{equation}
     \label{eqn:fluctuation}
 Score_{Fluc} = \max(\underset{1 \leq i \leq m_A}{\max}(d(CA_i,CA_{i+1})), d(CA_{m_A}, CB_1)).
\end{equation}
where $d(CA_i, CA_{i+1})$ is also called \textbf{adjacent distance} of node $CA_{i}$, i.e, the distance between a node and its previously obtained node in the sub-chain, and $d(CA_{m_A}, CB_1)$ is the distance between the junction nodes.  
\vspace{-2mm}
\subsubsection*{Evolving Score} Next, we would expect that $JTSC_{AB}$ represents some obvious evolving trend in $T_B$ whose later behavior is divergent from its earlier behavior in $T_B$. Thus, the distance between the start and end nodes of the chain in the target time series $T_B$ is used to evaluate such evolving capability. Specifically,
\vspace{-1mm}
\begin{equation}
    \label{eqn:evolv}
    Score_{Evolve} = d(CB_{1}, CB_{m_B}).
\end{equation}
A larger evolving Score means the chain has some obvious divergent behaviors.
\vspace{-2mm}
\subsubsection*{Reference Deviation Score}
In our problem, we are more interested in \textit{distinct evolving trend} that does not happen in $T_A$. If a detected chain node in $T_B$ is similar to some subsequence in $T_A$, it may indicate some periodic activity in the system and hence less useful for our purpose. We propose a Reference Deviation Score ($Score_{Ref}$) based on AB/BA Join Matrix Profile to measure such behavior. $Score_{Ref}$ is given by:
\vspace{-1mm}
\begin{equation}
    \label{eqn:Score_RD}
    Score_{Ref} = Top^k_{1\leq j\leq m_B}(MP_{BA}(1, CB_{j})),
\end{equation}
where $Top^k(\cdot)$ is the $k^{th}$ largest nearest neighbor distance from BA Join Matrix Profile. Note that instead of using the maximum values, the $k^{th}$ largest distance keeps our ranking robust to extreme outliers and noises. A large Reference Deviance score indicates the chain is more unique and more likely to capture distinct evolving in $T_B$. 
\vspace{-2mm}
\subsubsection*{Adjacent-distance Based Node Filtering}
Similar with Imamura et al. \cite{imamura2020matrix} in TSC20, we conduct a node filtering step to further enhance the robustness of the obtained JTSCs against extreme outliers and noisy nodes. The quantile of AB/BA Join matrix profile (such as median) is used to determine the noisiness of the node. The sub-chain propagation will early-terminate when reaching a node whose adjacent distance is greater than the distance threshold determined by the above quantile. 

\vspace{-2mm}
\subsection*{Difference from the Ranking Score of TSC20}Although TSC20 has a ranking criterion as well, our ranking score is better designed for our problem, and more focused on finding \textit{unusual and distinct cross-time-series evolving patterns}. In comparison, our fluctuation score relaxes the constraint in $T_B$ to allow unusual chain detection. Moreover, our ranking criterion considers a reference deviation score,  enforcing chain nodes in $T_B$ to be dissimilar from \textit{all} subsequences in $T_A$, while TSC20 only considers similarity between the first and last nodes. As a result, they might not be able to identify a chain that represents unusual evolving trend with large deviation in the target time series $T_B$.

\vspace{-3mm}
\subsection{Algorithm and Complexity of Finding Top Ranked JTSC}
\begin{algorithm}[t]
    \caption{JTSC Discovery Algorithm}
 \begin{algorithmic}[1]
    \STATE \textbf{Input}: Time Series $T_A$, $T_B$, Subsequence length $l$
    \STATE \textbf{Output}: top ranking Chain $JTSC_{best}$ 
    
    \\{\color{blue} /* Compute Matrix Profiles*/}
    
    \STATE $MP^A_L=$LeftMatrixProfile($T_A$,$l$)
    
    \STATE $MP^B_R=$RightMatrixProfile($T_B$,$l$)
    
    \STATE $MP_{AB},MP_{BA}=$JoinMatrixProfile($T_A$,$T_B$,$l$)

    \\{\color{blue} /* Extract Backward/Forward Chain based on Def. 11 and 12*/}
    
    \STATE $Ch_A=$BackwardChain($MP^A_L$)
    
    \STATE $Ch_B=$ForwardChain($MP^B_R$)
     
     \\{\color{blue} /*Combine Chain based on Def. 13*/}
     
    \STATE $Ch_{J}=$CombineChain($Ch_A$,$Ch_B$,$MP_{AB}$,$MP_{BA}$)
    
    \\{\color{blue} /* Rank Top Candidates Sec. 6.1*/}
    \STATE $Ch_{J}=$NodeFilter($Ch_{J}$, $MP_{AB}$, $MP_{BA}$)
    
    \STATE $JTSC_{best}=$RankTopChain($Ch_{J}$)
    
    \STATE \textbf{return} $JTSC_{best}$
  \end{algorithmic}
\end{algorithm}

The overall algorithm is shown in Algorithm 1. Given two input time series $T_A$ and $T_B$, the algorithm first computes all required matrix profiles (Lines 3-7). Then, all joint time series chain candidates are extracted based on Definition 13 (Lines 8-11). Finally, we rank and identify the best chain based on Eq. 6.1 (Lines 13-14). Similar to TSC17 and TSC20, most of the time spent is in computing matrix profiles. Therefore, our time complexity is $O(n_A^2+n_B^2)$, which is competitive with all existing time series chain methods.
\vspace{-5mm}
\section{EMPIRICAL EVALUATION}
\subsection{Experiment Setup} We perform empirical evaluation by conducting both quantitative experiments and a case study to demonstrate the interpretability and the utility of our method in a real-world scenario. We first describe the experimental setup on our proposed method and the baseline methods. 
\vspace{-2mm}
\subsubsection{Baseline Methods} We compare the JTSC algorithm with two state-of-the-art methods: 
\begin{itemize}
\vspace{-2mm}
    \item \textbf{TSC17} The bi-directional time series chain proposed by Zhu et al. \cite{zhu2017tsc}. 
    \vspace{-1mm}
    \item \textbf{TSC20} The backward time series chain proposed by Imamura et al. \cite{imamura2020matrix}.
    \vspace{-1mm}
\end{itemize}
Since our work is the first paper to consider the problem of detecting joint time series chain, and both TSC17 and TSC20 require running on a single time series, we run their algorithms on the concatenated time series from the reference and the target time series. 
\vspace{-2mm}
\subsubsection{Datasets}
\begin{figure}[ht]
    \centering
    \includegraphics[width=80mm]{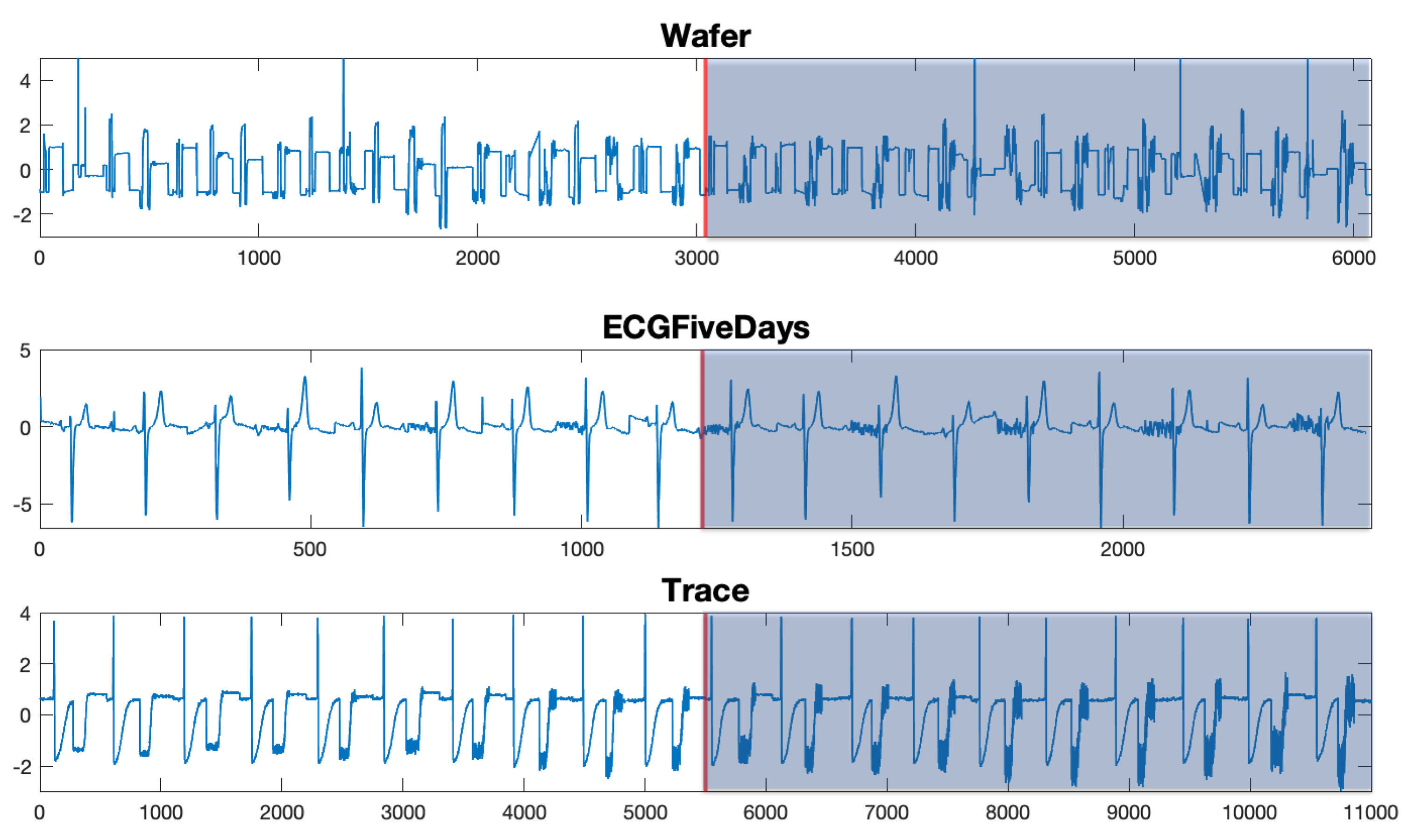}
    \vspace{-3mm}
    \caption{Samples of generated $T_A$ and $T_B$ from UCR data instances. The blue shaded portion is the target time series $T_B$.}
    \label{fig:data_gen}
    \vspace{-5mm}
\end{figure}
Since JTSC is a new problem, there is no suitable dataset with ground truth for quantitative analysis. To evaluate our method, we use 17 datasets from the widely-used UCR time series classification archive (https://www.cs.ucr.edu/$\sim$eamonn/time\_series\_data/) to generate time series covering different application domains where evolving trends are often observed. We use all sensor and ECG datasets with instance length less than 500, and device data instance length less than 750 to include more device datasets. Similar to our baselines \cite{zhu2017tsc, imamura2020matrix}, our method only focuses on detecting shape-based evolving trend in the time domain; therefore, we exclude datasets with information in frequency domain, or those with high amount of high frequency noises (e.g., Earthquake and FordA). We also exclude datasets that have extremely small numbers of instances, or contain randomly missing values.

For each dataset, 20 instances per class are randomly selected from two of the classes. If the data has slightly fewer than 20 instances per class, we use all of the instances. We form two time series, $T_A$ and $T_B$, which are generated by concatenating 10 z-normalized instances from each class alternately per class. Then, for the instances from a particular class in $T_A$, we inject uniform noise with linearly increasing levels from $\pm0.05$ to $\pm0.5$ to the first class instances only. Similarly, for $T_B$, we inject similar noise increasing from $\pm0.5$ to $\pm1$ to simulate the evolution of the sensor failure process to the instances in the same class. The length of noise is half of the instance length in the original classification dataset. We repeat this process 10 times per dataset.

Figure \ref{fig:data_gen} shows some sample plots of the time series we generated. The shaded time series is the Target time series $T_B$. It is easy to see that $T_A$ (Reference time series) is more stable. We can also see visually from the plots that the fluctuated noises increase overtime and in $T_B$, they become less stable. However, the overall shape is still maintained. The goal is to test whether our proposed method can detect such unexpected evolving patterns in time series. 

We also collected data from the actual production from Intel to perform our case study. We will provide more details on the data in Section 7.5. 
\vspace{-2mm}
\subsection{Parameter Setting}
To ensure fair comparison, we use all default parameters from both TSC17 and TSC20 and their original source code in Matlab. We use the length of the injected noises as the subsequence length, which is half of the instance length in the classification data. 

For our own method, we fix a single set of parameters throughout the experiments. The Chain threshold is set to be the quantile of 0.5 and 0.6 of the Join Matrix Profile. We use the $3^{rd}$ largest distance in our reference deviation score throughout all experiments. The chain with the highest overall ranking score will be reported.
\vspace{-4mm}
\subsection{Performance Measurement}
Inspired by Zhu et al.~\cite{zhu2017tsc}, we use the average hit rate to measure the performance of unusual trend captured in $T_B$. If a detected subsequence (i.e. a node in the chain) overlaps its closest ground truth by more than 25\% of the length of ground truth, we would count it as a hit. Then the overall hit rate is: 
\vspace{-2mm}
$$\text{Hit Rate} = \frac{\text{Number of Nodes Hits}}{\text{Total Number of Nodes}}.$$ 
The average hit rate from 10 time
series generated from each dataset is reported.

The range of hit-rate is between 0 and 1, where a higher hit rate means better performance. As discussed in Zhu et al. \cite{zhu2017tsc}, we are not aiming at detecting \textit{all} ground truth evolving trend (hit$\_$rate=1). For many applications, even observing a partial evolving trend could provide some guidance to the domain experts and potentially lead to some findings.

The hit rate provides more direct quantitative measure on the quality of detection instead of simply setting a threshold on hit or miss. Moreover, the hit rate allows effective measure on the evolving activity rather than how perfectly aligned the chains are from the ground truth.
\vspace{-4mm}
\subsection{Results}
\begin{table}[ht]
\vspace{-4mm}
\caption{Hit Rate for Synthetic Data consisting of sensor, ECG and device}
\scalebox{0.72}{
\begin{tabular}{@{}llcccc@{}}
\toprule
Dataset                & Type & $l_{instance}$ & TSC17         & TSC20 & JTSC          \\ \midrule
TwoLeadECG             & ECG       & 82              & 0.15          & 0.07  & \textbf{0.27} \\
ECG200                 & ECG       & 96              & 0.15          & 0.03  & \textbf{0.21} \\
ECGFiveDays            & ECG       & 136             & 0.22          & 0.1   & \textbf{0.36} \\
ECG5000                & ECG       & 140             & 0.15          & 0.01  & \textbf{0.21} \\
SmallKitchenAppliances & Device    & 720             & 0.08          & 0.02  & \textbf{0.19} \\
ScreenType             & Device    & 720             & 0.12          & 0.05  & \textbf{0.22} \\
PowerCons              & Device    & 144             & \textbf{0.13} & 0.04  & 0.11          \\
LargeKitchenAppliances & Device    & 720             & 0.04          & 0.02  & \textbf{0.17} \\
ElectricDevices        & Device    & 96              & 0.11          & 0.07  & \textbf{0.26} \\
Computers              & Device    & 720             & 0.05          & 0.05  & \textbf{0.13} \\
Lightning7             & Sensors   & 319             & 0.22          & 0.08  & \textbf{0.30} \\
FreezerRegularTrain    & Sensors   & 301             & 0.06          & 0     & \textbf{0.20} \\
Trace                  & Sensors   & 275             & 0.25          & 0.07  & \textbf{0.42} \\
Wafer                  & Sensors   & 152             & 0.05          & 0     & \textbf{0.30} \\
Plane                  & Sensors   & 144             & 0.17          & 0.06  & \textbf{0.37} \\
SonyAIBORobotSurface1  & Sensors   & 70              & \textbf{0.60} & 0.2   & 0.43          \\
SonyAIBORobotSurface2  & Sensors   & 65              & 0.20          & 0.08  & \textbf{0.37} \\ \hline
Total Wins             & -          & -                & 2             & 0     & \textbf{15}   \\ \bottomrule
\end{tabular}}
\end{table}

Table 1 shows the average hit rate of 10 runs for all the methods on the generated datasets from UCR archive. We use $l_{instance}$ to denote the original instance length from the data. The length of $T_A$ and $T_B$ are both $20\times l_{instances}$. Our proposed approach outperforms the baselines on 15 out of 17 datasets. The results show that the proposed method could better capture the evolving patterns than the current time series chain discovery methods across two time series. 

\vspace{-2mm}
\subsection{Case Study - Application in Intel Production Data}
\begin{figure}[h]
    \centering
    \includegraphics[width=75mm]{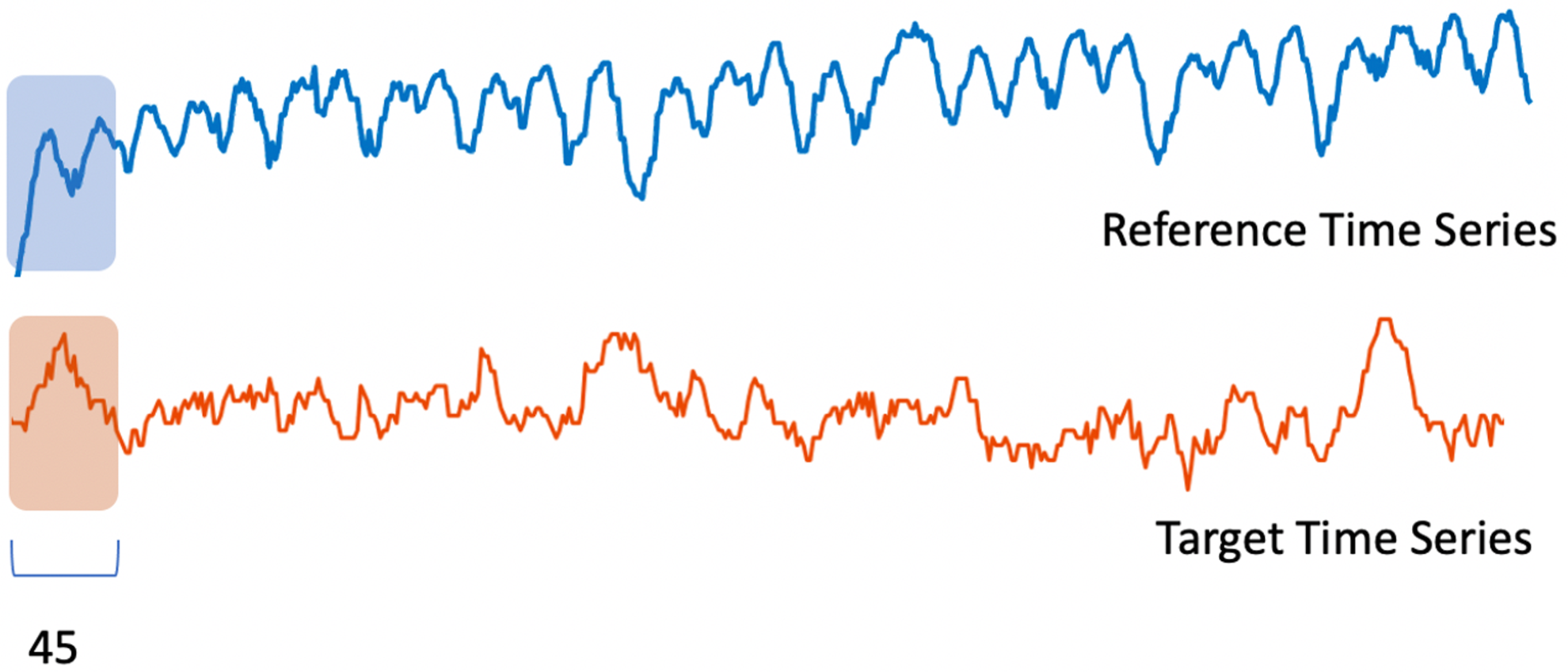}
    \vspace{-2mm}
    \caption{Two snippets from the reference and target time series from Intel Production Data.}
    \label{fig:case_study_overall}
    \vspace{-2mm}
\end{figure}
\vspace{-1mm}
\begin{figure}[h]
    \centering
    \includegraphics[width=80mm]{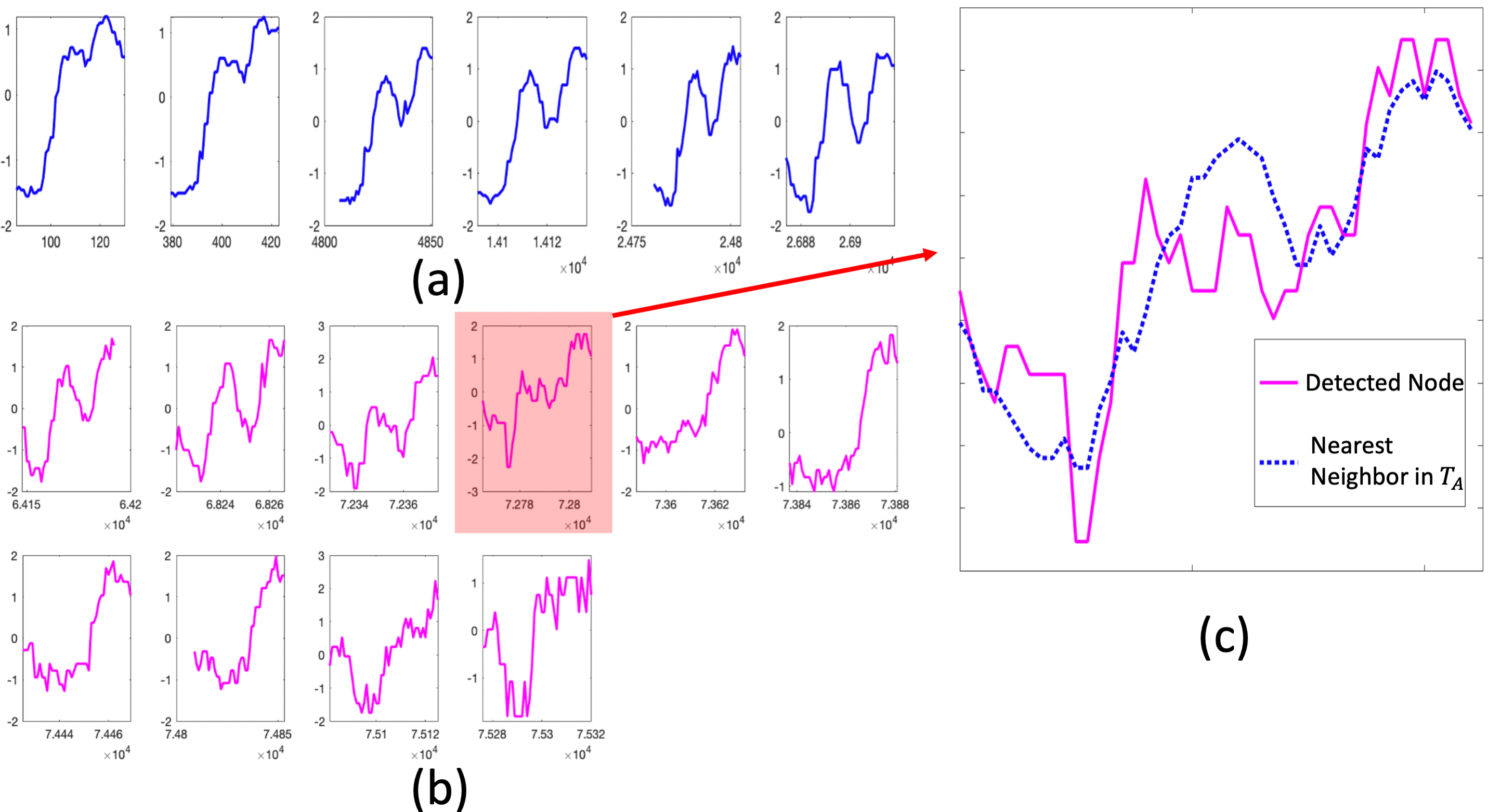}
    \vspace{-3mm}
    \caption{(a) Z-normalized reference sub-chain and (b) target sub-chain for Intel Production Data. (c) Zoom in view of the most deviance node in detected target sub-chain against its nearest neighbor in reference time series $T_A$. }
    \label{fig:Case_Study}
    \vspace{-5mm}
\end{figure}
\vspace{-1mm}
We apply our method to detect changes in behavior of a solder reflow oven used in the package assembly process at Intel Corporation. During solder reflow, the part is heated to a high temperature to melt the solder and then cooled down to room temperature before it leaves the oven. During this process, flux used to help the solder adhere to the pads evaporates, and during the cool down stage, this flux deposits within the oven. Excessive build up of flux would result in it dripping onto the parts resulting in a scrap event. As trays containing parts move through an oven zone, they perturb the temperature which causes the zone temperature controller to respond to pull it back to its setpoint. As flux builds up, we expect the thermal dynamics of the sensor and zone to change which should be detectable in the observed temperature time series trace. The method developed is applied to see if such a change can be detected which would allow (similar to the welding case mentioned in the Introduction) for preemptive maintenance that would result in the flux being cleaned off the interior of the oven.

The oven temperature is collected at a 0.5Hz sample rate when the oven is processing product and the overall time series we are looking at has 75756 data points. This time series is divided into two groups - series A which contains 38710 points and corresponds to the period after maintenance, and series B containing 37046 points which corresponds to the time series before a flux drip event was recorded on the oven. Figure \ref{fig:case_study_overall} shows a snippet of of length 450 time series from reference time series (blue) and target time series (red) respectively. From visual inspection, there is no obvious time series chain or obvious anomaly. We have to rely on a pattern detection algorithm to detect any variation. 

Figure~\ref{fig:Case_Study} shows the top joint time series chain found by our algorithm. Figure~\ref{fig:Case_Study} (a) and (b) are reference and target sub-chain for our top-1 ranked joint time series chain respectively. The goal is to monitor the system and also capture any unexpected dynamic response of the oven. As we are interested in the temperature fluctuation as a tray enters and exits the zone, we compute the expected time interval and using this to set our subsequence length to 45 data points. The result is very promising. Our algorithm is able to detect 6 nodes in reference sub-chain and 10 nodes in target sub-chain. We observe that the overall trend of the shape for pattern in reference sub-chain is relatively stable, while target sub-chain contain nodes with increasing fluctuation and perturbance. 

We also identify a potential flux drip event by noticing the unexpected behavior starting with the 4th node in target sub-chain. The pattern occurs about 25 minutes before a flux drip event occur. After Node 4 in target sub-chain, the temperature control starts showing more fluctuation. The temperature recovery becomes less smooth compared to pattern of nodes in $T_A$, reflecting that the oven has seen a change in thermal characteristics due to flux buildup. 

To further validate this, we check to see whether time series A has any similar pattern. We queried the nodes in target sub-chain to get their closest subsequence in the reference time series and compute the distance. Figure~\ref{fig:Case_Study} (c) shows the top discord node in target sub-chain with its closest neighbor in $T_A$. Compared to its nearest neighbor subsequence, the pattern for the top discord node has a lot more perturbations and does not follow the typical shape in any of reference sub-chain. Its nearest neighbor in $T_A$ is significantly more smooth in shape, also indicating the system is less stable and needs to be looked at. 

In summary, the detected joint time series chain reveals unexpected evolution in time series $T_B$, allowing a domain expert to interpret the results with domain knowledge and closely monitor a continuous process for dynamic change of the status in the oven. The evolving trend extracted is highly interpretable and is useful for providing insight and guidance for the engineers and practitioners in the manufacture process. As the ubiquitous existence of industrial control applications to regulate temperature, flow, pressure, speed and other process, time series chain could benefit more domain applications on detecting potential unusual evolving in offset and perturbances in many other domains and inspire more future research and industry applications broadly. 
\vspace{-3mm}
\section{CONCLUSION \& FUTURE WORK}
\vspace{-1mm}
 In conclusion, we introduce a new problem and a novel definition called Joint Time Series Chain (JTSC) to capture unexpected evolving trend across two time series with enhanced quality and robustness. We propose a new ranking criterion to effectively identify the most meaningful joint time series chain. Through the experiments, we demonstrate that JTSC outperforms the state-of-the-art methods, and is well suited for capturing unusual evolving trend in the data. In addition, we show the utility of JTSC by applying it to an Intel production data. Our result shows that JTSC reveals unusual evolution in sensor time series leading to the discovery of a flux dripping event. JTSC can potentially provide guidance for industrial practitioners and researchers broadly.  
 
 For future work, we would like to develop methods to identify joint TSC across multiple time series. 
 \vspace{-3mm}
\bibliography{TSC}
\vspace{-2mm}
\bibliographystyle{abbrv}
\end{document}